\documentclass[11pt,a4paper]{article}
\usepackage[hyperref]{acl2018}
\usepackage{times}
\usepackage{latexsym}
\usepackage{multirow}
\usepackage{bm}

\usepackage{url}
\usepackage{graphicx}
\usepackage{algorithm}
\usepackage{algorithmic}
\usepackage{mathrsfs}

\aclfinalcopy 
\def\aclpaperid{1469} 

\newenvironment{myitemize2}[1][]{
\begin{list}{$\bullet$}
    {
     \setlength{\leftmargin}{5mm}     
     \setlength{\parsep}{0.5mm}         
     \setlength{\topsep}{0mm}         
     \setlength{\itemsep}{0mm}        
     \setlength{\labelsep}{0.5em}     
     \setlength{\itemindent}{0mm}    
     \setlength{\listparindent}{6mm} 
    }}
{\end{list}}

\title{Multi-Granularity Hierarchical Attention Fusion Networks \\ for Reading Comprehension and Question Answering}
\author{Wei Wang\footnote{The authors contribute equally to this work.} , Ming Yan$^*$, Chen Wu$^*$ \\
{Alibaba Group, 969 West Wenyi Road, Hangzhou 311121, China} \\
{\tt \{hebian.ww,ym119608,wuchen.wc\}@alibaba-inc.com}
}

\begin{document}
\maketitle
\begin{abstract}
This paper describes a novel hierarchical attention network for reading comprehension style question answering, which aims to answer questions for a given narrative paragraph. In the proposed method, attention and fusion are conducted horizontally and vertically across layers at different levels of granularity between question and paragraph. Specifically, it first encode the question and paragraph with fine-grained language embeddings, to better capture the respective representations at semantic level. Then it proposes a multi-granularity fusion approach to fully fuse information from both global and attended representations. Finally, it introduces a hierarchical attention network to focuses on the answer span progressively with multi-level soft-alignment. Extensive experiments on the large-scale SQuAD and TriviaQA datasets validate the effectiveness of the proposed method. At the time of writing the paper (Jan. 12th 2018), our model achieves the first position on the SQuAD leaderboard for both single and ensemble models. We also achieves state-of-the-art results on TriviaQA, AddSent and AddOneSent datasets.
\end{abstract}



\section{Introduction}

As a brand new field in question answering community, reading comprehension is one of the key problems in artificial intelligence, which aims to read and comprehend a given text, and then answer questions based on it. This task is challenging which requires a comprehensive understanding of natural languages and the ability to do further inference and reasoning. Restricted by the limited volume of the annotated dataset, early studies mainly rely on a pipeline of NLP models to complete this task, such as semantic parsing and linguistic annotation~\cite{das2014frame}. Not until the release of large-scale cloze-style dataset, such as Children's Book Test~\cite{hill2015goldilocks} and CNN/Daily Mail~\cite{hermann2015teaching}, some preliminary end-to-end deep learning methods have begun to bloom and achieve superior results in reading comprehension task~\cite{hermann2015teaching,chen2016thorough,cui2016attention}. 

However, these cloze-style datasets still have their limitations, where the goal is to predict the single missing word (often a named entity) in a passage. It requires less reasoning than previously thought and no need to comprehend the whole passage~\cite{chen2016thorough}. Therefore, Stanford publish a new large-scale dataset SQuAD~\cite{rajpurkar2016squad}, in which all the question and answers are manually created through crowdsourcing. Different from cloze-style reading comprehension dataset, SQuAD constrains answers to all possible text spans within the reference passage, which requires more logical reasoning and content understanding. 

Benefiting from the availability of SQuAD benchmark dataset, rapid progress has been made these years. The work \cite{wang2016machine} and \cite{seo2016bidirectional} are among the first to investigate into this dataset, where Wang and Jiang propose an end-to-end architecture based on match-LSTM and pointer networks~\cite{wang2016machine}, and Seo et al. introduce the bi-directional attention flow network which captures the question-document context at different levels of granularity~\cite{seo2016bidirectional}. Chen et al. devise a simple and effective document reader, by introducing a bilinear match function and a few manual features~\cite{chen2017reading}. Wang et al. propose a gated attention-based recurrent network where self-match attention mechanism is first incorporated~\cite{wang2017gated}. In \cite{liu2017stochastic} and \cite{shen2017reasonet}, the multi-turn memory networks are designed to simulate multi-step reasoning in machine reading comprehension.

The idea of our approach derives from the normal human reading pattern. First, people scan through the whole passage to catch a glimpse of the main body of the passage. Then with the question in mind, people make connection between passage and question, and understand the main intent of the question related with the passage theme. A rough answer span is then located from the passage and the attention can be focused on to the located context. Finally, to prevent from forgetting the question, people come back to the question and select a best answer according to the previously located answer span.

Inspired by this, we propose a hierarchical attention network which can gradually focus the attention on the right part of the answer boundary, while capturing the relation between the question and passage at different levels of granularity, as illustrated in Figure~\ref{fig:frame}. Our model mainly consists of three joint layers: 1) encoder layer where pre-trained language models and recurrent neural networks are used to build representation for questions and passages separately; 2) attention layer in which hierarchical attention networks are designed to capture the relation between question and passage at different levels of granularity; 3) match layer where refined question and passage are matched under a pointer-network~\cite{vinyals2015pointer} answer boundary predictor.

In encoder layer, to better represent the questions and passages in multiple aspects, we combine two different embeddings to give the fundamental word representations. In addition to the typical glove word embeddings, we also utilize the ELMo embeddings~\cite{peters2018deep} derived from a pre-trained language model, which shows superior performance in a wide range of NLP problems. Different from the original fusion way for intermediate layer representations, we design a representation-aware fusion method to compute the output ELMo embeddings and the context information is also incorporated by further passing through a bi-directional LSTM network.

The key in machine reading comprehension solution lies in how to incorporate the question context into the paragraph, in which attention mechanism is most widely used. Recently, many different attention functions and types have been designed~\cite{xiong2016dynamic,seo2016bidirectional,wang2017gated}, which aims at properly aligning the question and passage. In our attention layer, we propose a hierarchical attention network by leveraging both the co-attention and self-attention mechanism, to gradually focus our attention on the best answer span. Different from the previous attention-based methods, we constantly complement the aligned representations with global information from the previous layer, and an additional fusion layer is used to further refine the representations. In this way, our model can make some minor adjustment so that the attention will always be on the right place.

Based on the refined question and passage representation, a bilinear match layer is finally used to identify the best answer span with respect to the question. Following the work of~\cite{wang2016machine}, we predict the start and end boundary within a pointer-network output layer. 

The proposed method achieves state-of-the-art results against strong baselines. Our single model achieves 79.2\% EM and 86.6\% F1 score on the hidden test set, while the ensemble model further boosts the performance to 82.4\% EM and 88.6\% F1 score. At the time of writing the paper (Jan. 12th 2018), our model SLQA+ (Semantic Learning for Question Answering) achieves the first position on the SQuAD leaderboard~\footnote{\small{~https://rajpurkar.github.io/SQuAD-explorer/}} for both single and ensemble models. Besides, we are also among the first to surpass human EM performance on this golden benchmark dataset.

\section{Related Work}
\subsection{Machine Reading Comprehension}
Traditional reading comprehension style question answering systems rely on a pipeline of NLP models, which make heavy use of linguistic annotation, structured world knowledge, semantic parsing and similar NLP pipeline outputs~\cite{hermann2015teaching}. Recently, the rapid progress of machine reading comprehension has largely benefited from the availability of large-scale benchmark datasets and it is possible to train large end-to-end neural network models. Among them, CNN/Daily Mail~\cite{hermann2015teaching} and Children's Book Test~\cite{hill2015goldilocks} are the first large-scale datasets for  reading comprehension task. However, these datasets are in cloze-style, in which the goal is to predict the missing word (often a named entity) in a passage. Moreover, Chen at al. have also shown that these cloze-style datasets requires less reasoning than previously thought~\cite{chen2016thorough}. Different from the previous datasets, the SQuAD provides a more challenging benchmark dataset, where the goal is to extract an arbitrary answer span from the original passage. 

\subsection{Attention-based Neural Networks}
The key in MRC task lies in how to incorporate the question context into the paragraph, in which attention mechanism is most widely used. In spite of a variety of model structures and attention types~\cite{cui2016attention,xiong2016dynamic,seo2016bidirectional,wang2017gated,clark2017simple}, a typical attention-based neural network model for MRC first encodes the symbolic representation of the question and passage in an embedding space, then identify answers with particular attention functions in that space. In terms of the question and passage attention or matching strategy, we roughly categorize these attention-based models into two large groups: one-way attention and two-way attention. 

In one-way attention model, question is first summarized into a single vector and then directly matched with the passage. Most of the end-to-end neural network methods on the cloze-style datasets are based on this model~\cite{hermann2015teaching,kadlec2016text,chen2016thorough,dhingra2016gated}. Hermann et al. are the first to apply the attention-based neural network methods to MRC task and introduce an attentive reader and an impatient reader~\cite{hermann2015teaching}, by leveraging a two layer LSTM network. Chen et al.~\cite{chen2016thorough} further design a bilinear attention function based on the attentive reader, which shows superior performance on CNN/Daily Mail dataset. However, part of information may be lost when summarizing the question and a fine-grained attention on both the question and passage words should be more reasonable. 

Therefore, the two-way attention model unfolds both the question and passage into respective word embeddings, and compute the attention in a two-dimensional matrix. Most of the top-ranking methods on SQuAD leaderboard are based on this attention mechanism~\cite{wang2017gated,huang2017fusionnet,xiong2017dcn+,liu2017stochastic,liu2017phase}. \cite{cui2016attention} and \cite{xiong2016dynamic} introduce the co-attention mechanism to better couple the representations of the question and document. Seo et al. propose a bi-directional attention flow network to capture the relevance at different levels of granularity~\cite{seo2016bidirectional}. \cite{wang2017gated} further introduce the self-attention mechanism to refine the representation by matching the passage against itself, to better capture the global passage information. Huang et al. introduce a fully-aware attention mechanism with a novel \emph{history-of-word} concept~\cite{huang2017fusionnet}.

We propose a hierarchical attention network by leveraging both co-attention and self-attention mechanisms in different layers, which can capture the relevance between the question and passage at different levels of granularity. Different from the above methods, we further devise a fusion function to combine both the aligned representation and the original representation from the previous layer within each attention. In this way, the model can always focus on the right part of the passage, while keeping the global passage topic in mind.



\section{Machine Comprehension Model}

\subsection{Task Description}
Typical machine comprehension systems take an evidence text and a question as input, and predict a span within the evidence that answers the question. Based on this definition, given a passage and a question, the machine needs to first read and understand the passage, and then finds the answer to the question. The passage is described as a sequence of word tokens $\rm P=\left \{ w^P_t\right \}^n_{t=1}$ and the question is described as $\rm Q=\left \{ w^Q_t\right \}^m_{t=1}$, where $\rm n$ is the number of words in the passage, and $\rm m$ is the number of words in the question. In general, $\rm n\gg m$. The answer can have different types depending on the task. In the SQuAD dataset~\cite{rajpurkar2016squad}, the answer $\rm A$ is guaranteed to be a continuous span in the passage $\rm P$. The object function for machine reading comprehension is to learn a function $\rm f(q,p)=\arg\max_{a \in A(p)} P(a|q,p)$. The training data is a set of the question, passage and answer tuples $\rm <Q, P, A>$.

\subsection{Encode-Interaction-Pointer Framework}
\begin{figure}
\centering
\includegraphics[width=0.495\textwidth]{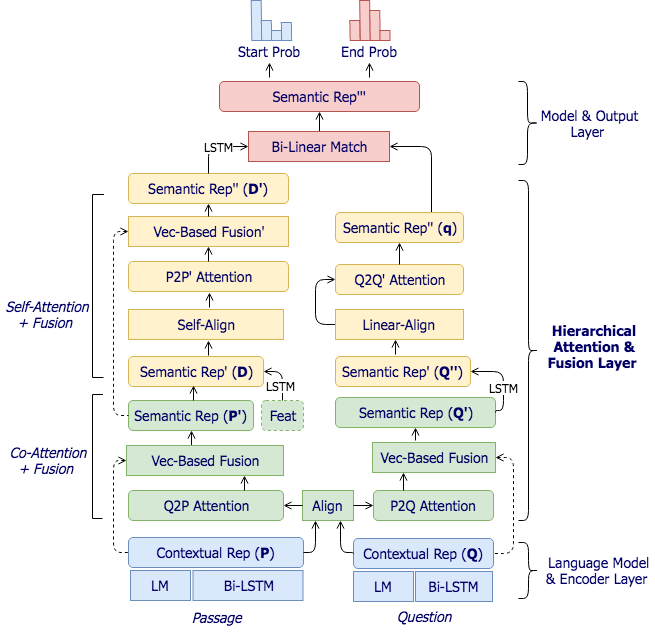}\vspace{-1mm}
\caption{Hierarchical Attention Fusion Network.}
\label{fig:frame}
\end{figure}

We will now describe our framework from the bottom up. As show in Figure~\ref{fig:frame}, the proposed framework consists of four typical layers to learn different concepts of semantic representations: 
\vspace{2mm}

\begin{myitemize2}
  \item \textbf{Encoder Layer} as a language model, utilizes contextual cues from surrounding words to refine the embedding of the words. It converts the passage and question from tokens to semantic representation;
  \item \textbf{Attention Layer} attempts to capture relations between question and passage. Besides the aligned context, the contextual embeddings are also merged by a fusion function. Moreover, the multi-level of this operation forms a "working memory";
  \item \textbf{Match Layer} employs a bi-linear match function to compute the relevance between the question and passage representation on a span level;
  \item  \textbf{Output Layer} uses a pointer network to search the answer span of question.
\end{myitemize2}





The main contribution of this work is the attention layer, in order to capture the relationship between question and passage, a hierarchical strategy is used to progressively make the answer boundary clear with the refined attention mechanism.  A fine-grained fusion function is also introduced to better align the contextual representations from different levels. The detailed description of the model is provided as follows.

\subsection{Hierarchical Attention Fusion Network}
Our design is based on a simple but natural intuition: performing fine-grained mechanism requires first to roughly see the potential answer domain and then progressively locate the most discriminative parts of the domain. 

The overall framework of our Hierarchical Attention Fusion Network is shown in Figure~\ref{fig:frame}. It consists of several parts: a basic co-attention layer with shallow semantic fusion, a self-attention layer with deep semantic fusion and a memory-wise bilinear alignment function. The proposed network has two distinctive characteristics: (i) A fine-grained fusion approach to blend attention vectors for a better understanding of the relationship between question and passage; (ii) A multi-granularity attention mechanism applied at the word and sentence-level, enabling it to properly attend to the most important content when constructing the question and passage representation. Experiments conducted on SQuAD and adversarial example datasets~\cite{jia2017adversarial} demonstrate that the proposed framework outperform previous methods by a large margin. Details of different components will be described in the following sections.


\subsection{ Language Model \& Encoder Layer}
Encoder layer of the model transform the discrete word tokens of question and passage to a sequence of continuous vector representations. We use a pre-trained word embedding model and a char embedding model to lay the foundation for our model. For the word embedding model, we adopt the popular glove embeddings~\cite{pennington2014glove} which are widely used in deep learning-based NLP domain. For the char embedding model, the ELMo language model~\cite{peters2018deep} is used due to its superior performance in a wide range of NLP tasks. As a result, we obtain two types of encoding vectors, i.e., word embeddings $\rm \left \{ e^Q_t\right \}^m_{t=1}$,  $\rm \left \{ e^P_t\right \}^n_{t=1} $ and char embeddings  $\rm \left \{ c^Q_t\right \}^m_{t=1}$, $\rm \left \{ c^P_t\right \}^n_{t=1}$.


To further utilize contextual cues from surrounding words to refine the embedding of the words, we then put a shared Bi-LSTM network on top of the embeddings provided by the previous layers to model the temporal interactions between words. Before feeding into the Bi-LSTM contextual network, we concat the word embeddings and char embeddings for a full understanding of each word. The final output of our encoder layer is shown as below,
\begin{equation}\label{equ:2}
\rm u_t^Q = \left[ {BiLST{M_Q}([e_t^Q,c_t^Q]),c_t^Q} \right]
\end{equation}
\begin{equation}\label{equ:3}
\rm u_t^P = \left[ {BiLST{M_P}([e_t^P,c_t^P]),c_t^P} \right]
\end{equation}
where we further concat the output of the contextual Bi-LSTM network with the pre-trained char embeddings for its good performance~\cite{peters2018deep}. This can be regarded as a residual connection between word representations in different levels. 




\subsection{Hierarchical Attention \& Fusion Layer}
The attention layer is responsible for linking and fusing information from the question and passage representation, which is the most critical in most MRC tasks. It aims to align the question and passage so that we can better locate on the most relevant passage span with respect to the question. We propose a hierarchical attention structure by combining  the co-attention and self-attention mechanism in a multi-hop style. Besides, we think that the original representation and the aligned representation via attention can reflect the content semantics in different granularities. Therefore, we also apply a particular fusion function after each attention function, so that different levels of semantics can be better incorporated towards a better understanding.

\subsubsection{Co-attention \& Fusion} \label{sec:3}
Given the question and passage representation $\rm u^Q_t$ and $\rm u^P_t$, a soft-alignment matrix $\rm S$ has been built to calculate the shallow semantic similarity between question and passage as follows:
\begin{equation}\label{equ:4}
\small
\rm S_{ij}=Att( u^Q_t, u^P_t)=ReLU(W^\top_{lin} u^Q_t)^\top \cdot ReLU(W^\top_{lin} u^P_t)
\end{equation}
where $\rm W_{lin}$ is a trainable weight matrix.

This decomposition avoids the quadratic complexity that is trivially parallelizable~\cite{parikh2016decomposable}. Now we use the unnormalized attention weights $\rm S_{ij}$ to compute the attentions between question and passage, which is further used to obtain the attended vectors in passage to question and question to passage direction, respectively.

\textbf{P2Q Attention} signifies which question words are most relevant to each passage word, given as below:
\begin{equation}\label{equ:5}
\rm \textbf{$\alpha$}_j=softmax(S_{:j})
\end{equation}
where $\rm \alpha_j$ represents the attention weights on the question words.

The aligned passage representation from question $\rm Q= \left \{ u^Q_t\right \}^m_{t=1}$ can thus be derived as,
\begin{equation}\label{equ:6}
\rm \tilde{Q}_{:t}=\sum_{j}\alpha_{tj}\cdot Q_{:j}  ,\forall j\in [1,...,m]
\end{equation}
\textbf{Q2P Attention} signifies which passage words have the closest similarity to one of the question words and are hence critical for answering the question.

We utilize the same way to calculate this attention as in the passage to question attention (P2Q), except for that in the opposite direction:
\begin{equation}\label{equ:7}
\rm \textbf{$\beta$}_i=softmax(S_{i:})
\end{equation}
\begin{equation}\label{equ:8}
\rm \tilde{P}_{k:}=\sum_{i}\beta_{ik}\cdot P_{i:}  ,\forall i\in [1,...,n]
\end{equation}
where $\rm \tilde{P}$ indicates the weighted sum of the most important words in the passage with respect to the question.


With the aligned passage and question representations $\rm \tilde{Q}$ and $\rm \tilde{P}$ derived, a particular fusion unit has been designed to combine the original contextual representations and the corresponding attention vectors for question and passage separately:
\begin{equation} \label{equ:9} 
\rm {P}'=Fuse(P, \tilde{Q}) 
\end{equation}
\begin{equation} \label{equ:10} 
\rm {Q}'=Fuse(Q, \tilde{P})
\end{equation}
where $\rm Fuse(\cdot, \cdot)$ is a typical fusion kernel.

The simplest way of fusion is a concatenation or addition of the two representations, followed by some linear or non-linear transformation. Recently, a heuristic matching trick with difference and element-wise product is found effective in combining different representations~\cite{mou2016natural, chen2017natural}:
\begin{equation} \label{equ:30} 
\rm {m(P, \tilde{Q})}=tanh(W_{f}[P;\tilde{Q};P \circ \tilde{Q}; P-\tilde{Q}]+b_{f})
\end{equation}
where $\circ$ denotes the element-wise product, and $\rm W_{f}$, $\rm b_{f}$ are trainable parameters. The output dimension is projected back to the same size as the original representation $\rm P$ or $\rm Q$ via the projected matrix $\rm W_{f}$.

Since we find that the original contextual representations are important in reflecting the semantics at a more global level, we also introduce different levels of gating mechanism to incorporate the projected representations $\rm m(\cdot,\cdot)$ with the original contextual representations. As a result, the final fused representations of passage and question can be formulated as: 
\begin{equation} \label{equ:31} 
\rm {P}'=g(P, \tilde{Q}) \cdot {m(P, \tilde{Q})} + (1-g(P, \tilde{Q})) \cdot P
\end{equation}
\begin{equation} \label{equ:32} 
\rm {Q}'=g(Q, \tilde{P}) \cdot {m(Q, \tilde{P})} + (1-g(Q, \tilde{P})) \cdot Q
\end{equation}
where $\rm g(\cdot,\cdot)$ is a gating function. To capture the relation between the representations in different granularities, we also design a scalar-based, a vector-based and a matrix-based sigmoid gating function, which are compared in Section~\ref{sec:45}.


\subsubsection{Self-attention \& Fusion} 
Borrowing the idea from wide and deep network~\cite{cheng2016wide}, manual features have also been added to combine with the outputs of previous layer for a more comprehensive representation. In our model, these features are concatenated with the refined question-aware passage representation as below:
\begin{equation}\label{equ:11}
\rm D=BiLSTM([{P}';feat_{man}])
\end{equation}
where $\rm feat_{man}$ denotes the word-level manual passage features. 

In this layer, we separately consider the semantic representations of question and passage, and further refine the obtained information from the co-attention layer. Since fusing information among context words allows contextual information to flow close to the correct answer, the self-attention layer is used to further align the question and passage representation against itself, so as to keep the global sequence information in memory. Benefiting from the advantage of self-alignment attention in addressing the long-distance dependence~\cite{wang2017gated}, we adopt a self-alignment fusion process in this level. To allow for more freedom of the aligning process, we introduce a bilinear self-alignment attention function on the passage representation: 
\begin{equation}\label{equ:12}
\rm L=softmax(D\cdot W_l\cdot D^\top)
\end{equation}
\begin{equation}\label{equ:13}
\rm \tilde{D}=L \cdot D
\end{equation}

Another fusion function $\rm Fuse(\cdot, \cdot)$ is again adopted to combine the question-aware passage representation $\rm D$ and self-aware representation $\rm \tilde{D}$, as below:
\begin{equation}\label{equ:14}
\rm {D}'=Fuse(D, \tilde{D})
\end{equation}
 
Finally, a bidirectional LSTM is used to get the final contextual passage representation:
\begin{equation}\label{equ:15}
\rm {D}''=BiLSTM({D}')
\end{equation}
 
As for question side, since it is generally shorter in length and could be adequately represented with less information, we follow the question encoding method used in~\cite{chen2017reading} and adopt a linear transformation to encode the question representation to a single vector. 

First, another contextual bidirectional LSTM network is applied on top of the fused question representation: $\rm {Q}''=BiLSTM({Q}')$.
Then we aggregate the resulting hidden units into one single question vector, with a linear self-alignment: 
\begin{equation}\label{equ:17}
\rm \bm{\gamma}=softmax( \textbf{w}^\top_q\cdot {Q''})
\end{equation}
\begin{equation}\label{equ:18}
\rm \textbf{q}=\sum_{j}\gamma_j \cdot {Q}''_{:j}  ,\forall j\in [1,...,m]
\end{equation}
where $\rm \textbf{w}_q$ is a weight vector to learn, we self-align the refined question representation to a single vector according to the question self-attention weight, which can be further used to compute the matching with the passage words. 
 
\subsection{Model \& Output Layer}
Instead of predicting the start and end positions based only on $\rm {D}''$, a top-level bilinear match function is used to capture the semantic relation between question $\textbf{q}$ and paragraph $\rm {D}''$ in a matching style, which actually works as a multi-hop matching mechanism.

Different from the co-attention layer that generates coarse candidate answers and the self-attention layer that focus the relevant context of passage to a certain intent of question, the top model layer uses a bilinear matching function to capture the interaction between outputs from previous layers and finally locate on the right answer span. 

The start and end distribution of the passage words are calculated in a bilinear matching way as below,
\begin{eqnarray}\label{equ:19}
\rm P_{start}=softmax(\textbf{q}\cdot W_s^\top \cdot {D}'') \\
\rm P_{end}=softmax(\textbf{q}\cdot W_e^\top \cdot {D}'') 
\end{eqnarray}
where $\rm W_s$ and $\rm W_e$ are trainable matrices of the bilinear match function.

The output layer is application-specific, in MRC task, we use pointer networks to predict the start and end position of the answer, since it requires the model to find the sub-phrase of the passage to answer the question.

In training process, with cross entropy as metric, the loss for start and end position is the sum of the negative log probabilities of the true start and end indices by the predicted distributions, averaged over all examples:   
\begin{equation}\label{equ:21}
\rm L(\theta)=-\frac{1}{N}\sum_{i}^{N}\log p_{s}(y_{i}^{s})+\log p_{e}(y_{i}^{e})
\end{equation}
where $\rm \theta$ is the set of all trainable weights in the model,  and $\rm p_{s}$ is the probability of start index, $\rm p_{e}$ is the probability of end index, respectively. $\rm y_{i}^{s}$ and $\rm y_{i}^{e}$ are the true start and end indices.

During prediction, we choose the answer span with the maximum value of $\rm p_{s}\cdot p_{e}$ under a constraint that $\rm s \le e \le s + 15$, which is selected via a dynamic programming algorithm in linear time.

\section{Experiments}
In this section, we first present the datasets used for evaluation. Then we compare our end-to-end Hierarchical Attention Fusion Networks with existing machine reading models. Finally, we conduct experiments to validate the effectiveness of our proposed components. We evaluate our model on the task of question answering using recently released SQuAD and TriviaQA Wikipedia~\cite{joshi2017triviaqa}, which have gained a huge attention over the past year. An adversarial evaluation for the Stanford Question Answering SQuAD is also used to demonstrate the robust of our model under adversarial attacks~\cite{jia2017adversarial}. 

\subsection{Dataset}
We focus on the SQuAD dataset to train and evaluate our model. SQuAD is a popular machine comprehension dataset consisting of 100,000+ questions created by crowd workers on 536 Wikipedia articles. Each context is a paragraph from an article and the answer to each question is guaranteed to be a span in the context. The answer to each question is always a span in the context. The model is given a credit if its answer matches one of the human chosen answers. Two metrics are used to evaluate the model performance: Exact Match (EM) and a softer metric F1 score, which measures the weighted average of the precision and recall rate at a character level.

TriviaQA is a newly available machine comprehension dataset consisting of over 650K context-query-answer triples. The contexts are automatically generated from either Wikipedia or Web search results. The length of contexts in TriviaQA (average 2895 words) is much more longer than the one in SQuAD (average 122 words).

\subsection{Training Details}
We use the AdaMax optimizer, with a mini-batch size of 32 and initial learning rate of 0.002. A dropout rate of 0.4 is used for all LSTM layers. To directly optimize our target against the evaluation metrics, we further fine-tune the model with some well-defined strategy.  During fine-tuning, Focal Loss~\cite{lin2017focal} and Reinforce Loss which take F1 score as reward are incorporated with Cross Entropy Loss. The training process takes roughly 20 hours on a single Nvidia Tesla M40 GPU. We also train an ensemble model consisting of 15 training runs with the identical framework and hyper-parameters. At test time, we choose the answer with the highest sum of confidence scores amongst the 15 runs for each question.

\subsection{Main Results}
The results of our model and competing approaches on the hidden test set are summarized in Table ~\ref{tab:stat1}. The proposed SLQA+ ensemble model achieves an EM score of 82.4 and F1 score of 88.6, outperforming all previous approaches, which validates the effectiveness of our hierarchical attention and fusion network structure.

\begin{table}[t]
\scriptsize
\centering
\caption{\label{tab:stat1} The performance of our SLQA model and competing approaches on SQuAD.}
\begin{tabular}{l|c|c}
\hline
     & \textbf{Dev Set} & \textbf{Test Set} \\
\hline
\hline
    \emph{Single model}   & \textbf{EM / F1} & \textbf{EM / F1}  \\
    LR Baseline \cite{rajpurkar2016squad} & 40.0 / 51.0 & 40.4 / 51.0 \\
    Match-LSTM \cite{wang2016machine} &  64.1 / 73.9 & 64.7 / 73.7 \\
    DrQA \cite{chen2017reading} & - / - & 70.7 / 79.4 \\
    DCN+ \cite{xiong2017dcn+} & 74.5 / 83.1 & 75.1 / 83.1 \\
    Interactive AoA Reader+ \cite{cui2016attention} & - / - & 75.8 / 83.8 \\
    FusionNet \cite{huang2017fusionnet} & - / - & 76.0 / 83.9 \\
    SAN \cite{liu2017stochastic} & 76.2 / 84.0 & 76.8 / 84.4 \\
    AttentionReader+ (unpublished) & - / - & 77.3 / 84.9 \\
    BiDAF + Self Attention + ELMo \cite{peters2018deep} & - / - & 78.6 / 85.8 \\
    r-net+ \cite{wang2017gated} & - / - & 79.9 / 86.5 \\
    \textbf{SLQA+}  & \textbf{80.0 / 87.0} & \textbf{80.4 / 87.0}  \\
\hline
     \emph{Ensemble model}   &  &  \\
     FusionNet \cite{huang2017fusionnet} & - / - & 78.8 / 85.9 \\
     DCN+ \cite{xiong2017dcn+} & - / - & 78.9 / 86.0 \\
     Interactive AoA Reader+  \cite{cui2016attention} & - / - & 79.0 / 86.4 \\ 
     SAN \cite{liu2017stochastic} & 78.6 / 85.9 & 79.6 / 86.5 \\
     BiDAF + Self Attention + ELMo \cite{peters2018deep} & - / - & 81.0 / 87.4 \\
     AttentionReader+ (unpublished) & - / - & 81.8 / 88.2 \\
     r-net+ \cite{wang2017gated} & - / - & 82.6 / 88.5 \\
     \textbf{SLQA+}  & \textbf{82.0 / 88.4} & \textbf{82.4 / 88.6}  \\
\hline
Human Performance & 80.3 / 90.5 & 82.3 / 91.2 \\
\hline
\end{tabular}  \vspace{-3mm}
\end{table} 

We also conduct experiments on the adversarial SQuAD dataset~\cite{jia2017adversarial} to study the robustness of the proposed model. In the dataset, one or more sentences are appended to the original SQuAD context, aiming to mislead the trained models. We use exactly the same model as in our SQuAD dataset, the performance comparison result is shown in Table~\ref{tab:5}. It can be seen that the proposed model can still get superior results than all the other competing approaches.

\begin{table}[t]
\scriptsize
\centering
\caption{\label{tab:5} The F1 scores of different models on AddSent and AddOneSent datasets (S: Single Model, E: Ensemble).} 
\begin{tabular}{l|c|c}
\hline
    \emph{Model} & \textbf{AddSent} & \textbf{AddOneSent} \\
\hline
    Logistic \cite{rajpurkar2016squad} & 23.2 & 30.4 \\
    Match-S \cite{wang2016machine} &  27.3 & 39.0 \\
    Match-E  \cite{wang2016machine} &  29.4 & 41.8 \\
    BiDAF-S \cite{seo2016bidirectional} & 34.3 &  45.7 \\
    BiDAF-E \cite{seo2016bidirectional} & 34.2 &  46.9 \\
    ReasoNet-S \cite{shen2017reasonet} &  39.4 &  50.3 \\
    ReasoNet-E \cite{shen2017reasonet} &  39.4 &  49.8 \\
    Mnemonic-S \cite{hu2017reinforced} & 46.6 &  56.0 \\
    Mnemonic-E \cite{hu2017reinforced} & 46.2 &  55.3 \\
    QANet-S \cite{yu2018qanet} & 45.2 & 55.7  \\
    FusionNet-E \cite{huang2017fusionnet} & 51.4 & 60.7 \\
\hline
    \textbf{SLQA-S} (our) & \textbf{52.1} & \textbf{62.7} \\
    \textbf{SLQA-E} (our) & \textbf{54.8} & \textbf{64.2} \\
\hline
\end{tabular} \vspace{-3mm}
\end{table}




\subsection{Ablations}
In order to evaluate the individual contribution of each model component, we run an ablation study. Table ~\ref{tab:abl} shows the performance of our model and its ablations on SQuAD dev set. The bi-linear alignment plus fusion between passage and question is most critical to the performance on both metrics which results in a drop of nearly 15\%. The reason may be that in top-level attention layer, the similar semantics between question and passage are strong evidence to locate the correct answer span. The ELMo accounts for about 5\% of the performance degradation, which clearly shows the effectiveness of language model. We conjecture that language model layer efficiently encodes different types of syntactic and semantic information about words-in-context, and improves the task performance. To evaluate the performance of hierarchical architecture, we reduce the multi-hop fusion with the standard LSTM network. The result shows that multi-hop fusion outperforms the standard LSTM by nearly 5\% on both metrics.

\begin{table}[t]
\small
\centering
\caption{\label{tab:abl} Ablation tests of SLQA single model on the SQuAD dev set.}
\begin{tabular}{l|c}
\hline
    \emph{SLQA single model} & \textbf{EM} / \textbf{F1} \\
\hline
    \textbf{SLQA+} & \textbf{80.0 / 87.0} \\
    -Manual Features & 79.2 / 86.2\\
    -Language Embedding (ELMo) & 77.6 / 84.9\\
    -Self Matching & 79.5 / 86.4\\
    -Multi-hop & 79.1 / 86.1  \\
    -Bi-linear Match & 65.4 / 72.0 \\ 
    -Fusion (simple concat) & 78.8 / 85.8 \\
    -Fusion, -Multi-hop & 77.5 / 84.8 \\
    -Fusion, -Bi-linear Match & 63.1 / 69.6 \\
\hline
\end{tabular}  \vspace{-3mm}
\end{table}

\subsection{Fusion Functions} \label{sec:45}
In this section, we experimentally demonstrate how different choices of the fusion kernel impact the performance of our model. The compared fusion kernels are described as follows:

\textbf{Simple Concat}: a simple concatenation of two channel inputs.  

\textbf{Full Projection}: the heuristic matching and projecting function as in Equ.~\ref{equ:30}.

\textbf{Scalar-based Fusion}: the gating function is a trainable scalar parameter (a coarse fusion level): 
\begin{equation}\label{equ}
\rm g(P, \tilde{Q})=g_p
\end{equation}
where $\rm g_{p}$ is a trainable scalar parameter.

\textbf{Vector-based Fusion}: the gating function contains a weight vector to learn, which acts as a one-dimensional sigmoid gating,
\begin{equation}\label{equ}
\rm g(P, \tilde{Q}) = \sigma(\textbf{w}_{g}^\top \cdot [P;\tilde{Q};P \circ \tilde{Q}; P-\tilde{Q}]+b_{g})
\end{equation}
where $\rm \textbf{w}_{g}$ is trainable weight vector, $\rm b_{g}$ is trainable bias, and $\rm \sigma$ is sigmoid function.

\textbf{Matrix-based Fusion}: the gating function contains a weight matrix to learn, which acts as a two-dimensional sigmoid gating,
\begin{equation}\label{equ}
\rm g(P, \tilde{Q}) = \sigma(W_{g}^\top \cdot [P;\tilde{Q};P \circ \tilde{Q}; P-\tilde{Q}]+b_{g})
\end{equation}
where $\rm W_{g}$ is a trainable weight matrix.

The comparison results of different fusion kernels can be found in Table~\ref{tab:3}. We can see that different fusion methods contribute differently to the final performances, and the vector-based fusion method performs best, with a moderate parameter size.

\begin{table}[t]
\small
\centering
\caption{\label{tab:3} Comparison of different fusion kernels on the SQuAD dev set.}
\begin{tabular}{l|c}
\hline
    Fusion Kernel & \textbf{EM} / \textbf{F1} \\
\hline
    Simple Concat & 78.8 / 85.8 \\
    Add Full Projection (FPU) & 79.1 / 86.1\\
    Scalar-based Fusion (SFU) & 79.5 / 86.5\\
    \textbf{Vector-based Fusion (VFU)} & \textbf{80.0} / \textbf{87.0} \\
    Matrix-based Fusion (MFU) & 79.8 / 86.8  \\
\hline
\end{tabular}
\end{table}

\begin{table}[t]
\small
\centering
\caption{\label{tab:4} Comparison of different attention styles on the SQuAD dev set.}
\begin{tabular}{l|c}
\hline
    Attention Hierarchy & \textbf{EM} / \textbf{F1} \\
\hline
    1-layer attention (only  qp co-attention) & 61.9 / 68.4 \\
    2-layer attention (add self-attention) & 65.4 / 71.7 \\
    \textbf{3-layer attention (add bilinear match)} & \textbf{80.0} / \textbf{87.0}\\
\hline
    Attention Function & \textbf{EM} / \textbf{F1} \\
\hline
    dot product & 62.9 / 69.3 \\
    linear attention & 78.0 / 84.9 \\
    \textbf{bilinear attention (linear + relu)} & \textbf{80.0} / \textbf{87.0}  \\
    trilinear attention & 78.9 / 85.8  \\
\hline
\end{tabular}
\end{table}

\subsection{Attention Hierarchy and Function}
In the proposed model, attention layer is the most important part of the framework. At the bottom of Table~\ref{tab:4} we show the performances on SQuAD for four common attention functions. Empirically, we find bilinear attention which add ReLU after linearly transforming does significantly better than the others. 

At the top of Table~\ref{tab:4} we show the effect of varying the number of attention layers on the final performance. We see a steep and steady rise in accuracy as the number of layers is increased from N = 1 to 3. 



\begin{table}[t]
\scriptsize
\centering
\caption{\label{tab:6} Published and unpublished results on the TriviaQA wikipedia leaderboard.}
\begin{tabular}{l|c|c}
\hline
     & \textbf{Full} & \textbf{Verified} \\
\hline
   \emph{Model} & \textbf{EM} / \textbf{F1} & \textbf{EM} / \textbf{F1} \\
\hline
    BiDAF \cite{seo2016bidirectional} & 40.26 / 45.74 &  47.47 / 53.70 \\
    MEMEN \cite{pan2017memen} &  43.16 / 46.90 & 49.28 / 55.83 \\
    M-Reader \cite{hu2017reinforced} & 46.94 / 52.85 & 54.45 / 59.46 \\
    QANet \cite{yu2018qanet} &  51.10 / 56.60 &  53.30 / 59.20 \\
    document-qa \cite{clark2017simple} &  63.99 / 68.93 &  67.98 / 72.88 \\
    dirkweissenborn (unpublished) &  64.60 / 69.90 &  72.77 / 77.44 \\
\hline
    \textbf{SLQA-Single} & \textbf{66.56} / \textbf{71.39}  & \textbf{74.83} / \textbf{78.74} \\
\hline
\end{tabular} \vspace{-3mm}
\end{table} 

\subsection{Experiments on TriviaQA}
To further examine the robustness of the proposed model, we also test the model performance on TriviaQA dataset. The test performance of different methods on the leaderboard (on Jan. 12th 2018) is shown in Table~\ref{tab:6}. From the results, we can see that the proposed model can also obtain state-of-the-art performance in the more complex TriviaQA dataset.

\begin{figure}
\centering
\includegraphics[width=0.44\textwidth]{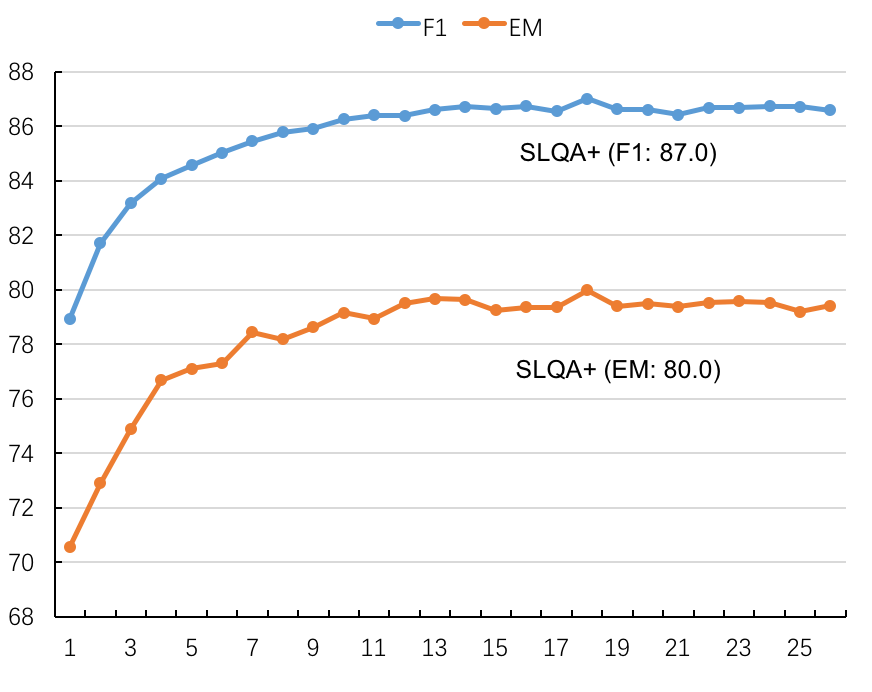} \vspace{-2mm}
\caption{Learning curve of F1 / EM score on the SQuAD dev set} \vspace{-3mm}
\label{fig:F1_EM}
\end{figure}


\section{Conclusions}
We introduce a novel hierarchical attention network, a state-of-the-art reading comprehension model which conducts attention and fusion horizontally and vertically across layers at different levels of granularity between question and paragraph. We show that our proposed method is very powerful and robust, which outperforms the previous state-of-the-art methods in various large-scale golden MRC datasets: SQuAD, TriviaQA, AddSent and AddOneSent. 


\section*{Acknowledgments}
We thank the Stanford NLP Group and the University of Washington NLP Group for evaluating our results on the SQuAD and the TriviaQA test set.

\bibliography{acl2018}

\begin{thebibliography}{32}
\expandafter\ifx\csname natexlab\endcsname\relax\def\natexlab#1{#1}\fi

\bibitem[{Chen et~al.(2016)Chen, Bolton, and Manning}]{chen2016thorough}
Danqi Chen, Jason Bolton, and Christopher~D Manning. 2016.
\newblock A thorough examination of the cnn/daily mail reading comprehension
  task.
\newblock \emph{arXiv preprint arXiv:1606.02858}.

\bibitem[{Chen et~al.(2017{\natexlab{a}})Chen, Fisch, Weston, and
  Bordes}]{chen2017reading}
Danqi Chen, Adam Fisch, Jason Weston, and Antoine Bordes. 2017{\natexlab{a}}.
\newblock Reading wikipedia to answer open-domain questions.
\newblock \emph{arXiv preprint arXiv:1704.00051}.

\bibitem[{Chen et~al.(2017{\natexlab{b}})Chen, Zhu, Ling, and
  Inkpen}]{chen2017natural}
Qian Chen, Xiaodan Zhu, Zhen-Hua Ling, and Diana Inkpen. 2017{\natexlab{b}}.
\newblock Natural language inference with external knowledge.
\newblock \emph{arXiv preprint arXiv:1711.04289}.

\bibitem[{Cheng et~al.(2016)Cheng, Koc, Harmsen, Shaked, Chandra, Aradhye,
  Anderson, Corrado, Chai, Ispir et~al.}]{cheng2016wide}
Heng-Tze Cheng, Levent Koc, Jeremiah Harmsen, Tal Shaked, Tushar Chandra,
  Hrishi Aradhye, Glen Anderson, Greg Corrado, Wei Chai, Mustafa Ispir, et~al.
  2016.
\newblock Wide \& deep learning for recommender systems.
\newblock In \emph{Proceedings of the 1st Workshop on Deep Learning for
  Recommender Systems}, pages 7--10. ACM.

\bibitem[{Clark and Gardner(2017)}]{clark2017simple}
Christopher Clark and Matt Gardner. 2017.
\newblock Simple and effective multi-paragraph reading comprehension.
\newblock \emph{arXiv preprint arXiv:1710.10723}.

\bibitem[{Cui et~al.(2016)Cui, Chen, Wei, Wang, Liu, and Hu}]{cui2016attention}
Yiming Cui, Zhipeng Chen, Si~Wei, Shijin Wang, Ting Liu, and Guoping Hu. 2016.
\newblock Attention-over-attention neural networks for reading comprehension.
\newblock \emph{arXiv preprint arXiv:1607.04423}.

\bibitem[{Das et~al.(2014)Das, Chen, Martins, Schneider, and
  Smith}]{das2014frame}
Dipanjan Das, Desai Chen, Andr{\'e}~FT Martins, Nathan Schneider, and Noah~A
  Smith. 2014.
\newblock Frame-semantic parsing.
\newblock \emph{Computational linguistics}, 40(1):9--56.

\bibitem[{Dhingra et~al.(2016)Dhingra, Liu, Yang, Cohen, and
  Salakhutdinov}]{dhingra2016gated}
Bhuwan Dhingra, Hanxiao Liu, Zhilin Yang, William~W Cohen, and Ruslan
  Salakhutdinov. 2016.
\newblock Gated-attention readers for text comprehension.
\newblock \emph{arXiv preprint arXiv:1606.01549}.

\bibitem[{Hermann et~al.(2015)Hermann, Kocisky, Grefenstette, Espeholt, Kay,
  Suleyman, and Blunsom}]{hermann2015teaching}
Karl~Moritz Hermann, Tomas Kocisky, Edward Grefenstette, Lasse Espeholt, Will
  Kay, Mustafa Suleyman, and Phil Blunsom. 2015.
\newblock Teaching machines to read and comprehend.
\newblock In \emph{Advances in Neural Information Processing Systems}, pages
  1693--1701.

\bibitem[{Hill et~al.(2015)Hill, Bordes, Chopra, and
  Weston}]{hill2015goldilocks}
Felix Hill, Antoine Bordes, Sumit Chopra, and Jason Weston. 2015.
\newblock The goldilocks principle: Reading children's books with explicit
  memory representations.
\newblock \emph{arXiv preprint arXiv:1511.02301}.

\bibitem[{Hu et~al.(2017)Hu, Peng, and Qiu}]{hu2017reinforced}
Minghao Hu, Yuxing Peng, and Xipeng Qiu. 2017.
\newblock Reinforced mnemonic reader for machine comprehension.
\newblock \emph{CoRR, abs/1705.02798}.

\bibitem[{Huang et~al.(2017)Huang, Zhu, Shen, and Chen}]{huang2017fusionnet}
Hsin-Yuan Huang, Chenguang Zhu, Yelong Shen, and Weizhu Chen. 2017.
\newblock Fusionnet: Fusing via fully-aware attention with application to
  machine comprehension.
\newblock \emph{arXiv preprint arXiv:1711.07341}.

\bibitem[{Jia and Liang(2017)}]{jia2017adversarial}
Robin Jia and Percy Liang. 2017.
\newblock Adversarial examples for evaluating reading comprehension systems.
\newblock In \emph{Proceedings of the 2017 Conference on Empirical Methods in
  Natural Language Processing}, pages 2021--2031.

\bibitem[{Joshi et~al.(2017)Joshi, Choi, Weld, and
  Zettlemoyer}]{joshi2017triviaqa}
Mandar Joshi, Eunsol Choi, Daniel~S Weld, and Luke Zettlemoyer. 2017.
\newblock Triviaqa: A large scale distantly supervised challenge dataset for
  reading comprehension.
\newblock \emph{arXiv preprint arXiv:1705.03551}.

\bibitem[{Kadlec et~al.(2016)Kadlec, Schmid, Bajgar, and
  Kleindienst}]{kadlec2016text}
Rudolf Kadlec, Martin Schmid, Ondrej Bajgar, and Jan Kleindienst. 2016.
\newblock Text understanding with the attention sum reader network.
\newblock \emph{arXiv preprint arXiv:1603.01547}.

\bibitem[{Lin et~al.(2017)Lin, Goyal, Girshick, He, and
  Doll{\'a}r}]{lin2017focal}
Tsung-Yi Lin, Priya Goyal, Ross Girshick, Kaiming He, and Piotr Doll{\'a}r.
  2017.
\newblock Focal loss for dense object detection.
\newblock \emph{arXiv preprint arXiv:1708.02002}.

\bibitem[{Liu et~al.(2017{\natexlab{a}})Liu, Wei, Mao, and
  Chikina}]{liu2017phase}
Rui Liu, Wei Wei, Weiguang Mao, and Maria Chikina. 2017{\natexlab{a}}.
\newblock Phase conductor on multi-layered attentions for machine
  comprehension.
\newblock \emph{arXiv preprint arXiv:1710.10504}.

\bibitem[{Liu et~al.(2017{\natexlab{b}})Liu, Shen, Duh, and
  Gao}]{liu2017stochastic}
Xiaodong Liu, Yelong Shen, Kevin Duh, and Jianfeng Gao. 2017{\natexlab{b}}.
\newblock Stochastic answer networks for machine reading comprehension.
\newblock \emph{arXiv preprint arXiv:1712.03556}.

\bibitem[{Mou et~al.(2016)Mou, Men, Li, Xu, Zhang, Yan, and
  Jin}]{mou2016natural}
Lili Mou, Rui Men, Ge~Li, Yan Xu, Lu~Zhang, Rui Yan, and Zhi Jin. 2016.
\newblock Natural language inference by tree-based convolution and heuristic
  matching.
\newblock In \emph{The 54th Annual Meeting of the Association for Computational
  Linguistics}, page 130.

\bibitem[{Pan et~al.(2017)Pan, Li, Zhao, Cao, Cai, and He}]{pan2017memen}
Boyuan Pan, Hao Li, Zhou Zhao, Bin Cao, Deng Cai, and Xiaofei He. 2017.
\newblock Memen: Multi-layer embedding with memory networks for machine
  comprehension.
\newblock \emph{arXiv preprint arXiv:1707.09098}.

\bibitem[{Parikh et~al.(2016)Parikh, T{\"a}ckstr{\"o}m, Das, and
  Uszkoreit}]{parikh2016decomposable}
Ankur Parikh, Oscar T{\"a}ckstr{\"o}m, Dipanjan Das, and Jakob Uszkoreit. 2016.
\newblock A decomposable attention model for natural language inference.
\newblock In \emph{Proceedings of the 2016 Conference on Empirical Methods in
  Natural Language Processing}, pages 2249--2255.

\bibitem[{Pennington et~al.(2014)Pennington, Socher, and
  Manning}]{pennington2014glove}
Jeffrey Pennington, Richard Socher, and Christopher Manning. 2014.
\newblock Glove: Global vectors for word representation.
\newblock In \emph{Proceedings of the 2014 conference on empirical methods in
  natural language processing (EMNLP)}, pages 1532--1543.

\bibitem[{Peters et~al.(2018)Peters, Neumann, Iyyer, Gardner, Clark, Lee, and
  Zettlemoyer}]{peters2018deep}
Matthew~E Peters, Mark Neumann, Mohit Iyyer, Matt Gardner, Christopher Clark,
  Kenton Lee, and Luke Zettlemoyer. 2018.
\newblock Deep contextualized word representations.
\newblock \emph{arXiv preprint arXiv:1802.05365}.

\bibitem[{Rajpurkar et~al.(2016)Rajpurkar, Zhang, Lopyrev, and
  Liang}]{rajpurkar2016squad}
Pranav Rajpurkar, Jian Zhang, Konstantin Lopyrev, and Percy Liang. 2016.
\newblock Squad: 100,000+ questions for machine comprehension of text.
\newblock \emph{arXiv preprint arXiv:1606.05250}.

\bibitem[{Seo et~al.(2016)Seo, Kembhavi, Farhadi, and
  Hajishirzi}]{seo2016bidirectional}
Minjoon Seo, Aniruddha Kembhavi, Ali Farhadi, and Hannaneh Hajishirzi. 2016.
\newblock Bidirectional attention flow for machine comprehension.
\newblock \emph{arXiv preprint arXiv:1611.01603}.

\bibitem[{Shen et~al.(2017)Shen, Huang, Gao, and Chen}]{shen2017reasonet}
Yelong Shen, Po-Sen Huang, Jianfeng Gao, and Weizhu Chen. 2017.
\newblock Reasonet: Learning to stop reading in machine comprehension.
\newblock In \emph{Proceedings of the 23rd ACM SIGKDD International Conference
  on Knowledge Discovery and Data Mining}, pages 1047--1055. ACM.

\bibitem[{Vinyals et~al.(2015)Vinyals, Fortunato, and
  Jaitly}]{vinyals2015pointer}
Oriol Vinyals, Meire Fortunato, and Navdeep Jaitly. 2015.
\newblock Pointer networks.
\newblock In \emph{Advances in Neural Information Processing Systems}, pages
  2692--2700.

\bibitem[{Wang and Jiang(2016)}]{wang2016machine}
Shuohang Wang and Jing Jiang. 2016.
\newblock Machine comprehension using match-lstm and answer pointer.
\newblock \emph{arXiv preprint arXiv:1608.07905}.

\bibitem[{Wang et~al.(2017)Wang, Yang, Wei, Chang, and Zhou}]{wang2017gated}
Wenhui Wang, Nan Yang, Furu Wei, Baobao Chang, and Ming Zhou. 2017.
\newblock Gated self-matching networks for reading comprehension and question
  answering.
\newblock In \emph{Proceedings of the 55th Annual Meeting of the Association
  for Computational Linguistics (Volume 1: Long Papers)}, volume~1, pages
  189--198.

\bibitem[{Xiong et~al.(2016)Xiong, Zhong, and Socher}]{xiong2016dynamic}
Caiming Xiong, Victor Zhong, and Richard Socher. 2016.
\newblock Dynamic coattention networks for question answering.
\newblock \emph{arXiv preprint arXiv:1611.01604}.

\bibitem[{Xiong et~al.(2017)Xiong, Zhong, and Socher}]{xiong2017dcn+}
Caiming Xiong, Victor Zhong, and Richard Socher. 2017.
\newblock Dcn+: Mixed objective and deep residual coattention for question
  answering.
\newblock \emph{arXiv preprint arXiv:1711.00106}.

\bibitem[{Yu et~al.(2018)Yu, Dohan, Luong, Zhao, Chen, Norouzi, and
  Le}]{yu2018qanet}
Adams~Wei Yu, David Dohan, Minh-Thang Luong, Rui Zhao, Kai Chen, Mohammad
  Norouzi, and Quoc~V Le. 2018.
\newblock Qanet: Combining local convolution with global self-attention for
  reading comprehension.
\newblock \emph{arXiv preprint arXiv:1804.09541}.

\end{thebibliography}
\bibliographystyle{acl_natbib}







\end{document}